\documentclass[runningheads]{llncs}
\usepackage{graphicx}
\usepackage{bbding}
\usepackage{multirow}

\begin{document}
\title{Multilingual Dialogue Generation with Shared-Private Memory}
\author{Chen Chen$^1$, Lisong Qiu$^2$ \and Zhenxin Fu$^2$ \and Dongyan Zhao$^2$ \and Junfei Liu$^{3}$ \and Rui Yan$^{2}$\thanks{Corresponding author: Rui Yan (ruiyan@pku.edu.cn).}}
\authorrunning{Chen et al.} % abbreviated author list (for running head)
%\tocauthor{Ronie Miguel Uliana}
%
\institute{$^1$ School of Software and Microelectronics of Peking University, Beijing, China,\\
%\email{chenchen@pku.edu.cn}\\
$^2$ Institute of Computer Science and Technology of Peking University, Beijing, China \\
$^3$ National Engineering Research Center for Software Engineering (Peking University), Beijing, China \\
\email{\{chenchen,qiulisong,fuzhenxin,zhaodongyan,liujunfei,ruiyan\}@pku.edu.cn}
}
\maketitle              % typeset the header of the contribution

\begin{abstract}
Existing dialog systems are all monolingual, where features shared among different languages are rarely explored.
In this paper, we introduce a novel multilingual dialogue system. Specifically, we augment the sequence to sequence framework with improved shared-private memory. 
The shared memory learns common features among different languages and facilitates a cross-lingual transfer to boost dialogue systems,
while the private memory is owned by each separate language to capture its unique feature. 
Experiments conducted on Chinese and English conversation corpora of different scales show that our proposed architecture outperforms the individually learned model with the help of the other language, where the improvement is particularly distinct when the training data is limited.
\end{abstract}

\section{Introduction}
Dialogue systems have long been an interest to the community of natural language processing due to their width range of applications.
These systems can be classified as task-oriented and non-task-oriented where task-oriented dialogue systems accomplish a specific task and non-task-oriented dialogue systems are designed to chat in open domain as chatbots  \cite{chen2017survey}.
In particular, the sequence-to-sequence ($\mathtt{Seq2Seq}$)
framework~\cite{sutskever2014sequence}, which learns to generate responses according to the given queries
%the mapping from the query to the response 
can achieve promising performance and grow popular~\cite{Serban2016Building}.
% Recent studies use generation-based methods by training a sequence-to-sequence neural network (\textit{Seq2Seq}) \cite{sutskever2014sequence} to build open-domain dialogue systems \cite{shang2015neural,sordoni2015a,Serban2016Building,li2016a,mei2016coherent}. These models directly generate responses according to the previous queries. 

Building a current state-of-the-art generation-based dialogue system requires large-scale conversational data. 
% However,  many languages are scarce resources, which makes these languages face challenges when using natural language processing techniques. Building a neural dialogue generation model requires large-scale training data due to numerous parameters that must be trained. 
However, the difficulty of collecting conversational data in different languages varies greatly~\cite{serban2015survey,li2017dailydialog}.
For example, it is difficult for minority languages to collect enough dialogue corpora to build a dialogue generation model as other majority languages (e.g., English and Chinese) do. 
%are more likely to collect enough dialogue corpora to build a better dialogue generation model than other minority languages.
Herein, we investigate to move the frontier of dialogue generation forward from a different angle.
More specifically, we find that some common features, e.g., dialogue logic, are shared in different languages but with different linguistic forms. 
%Some results of previous work also show that the multilingual system can perform better compared to its monolingual counterparts \cite{Mrk2017Semantic,liu2018proceedings}. 
Leveraging a multi-task framework for cross-lingual transfer learning can alleviate the problems caused by the scarcity of resources \cite{heigold2013multilingual,Dong2015Multi,2017arXiv170605137K}.
Through common dialogue features shared among different languages, the logic knowledge of different languages can be transferred and the robustness of the conversational model can be improved. 
However, to the best of our knowledge, no existing study has ever tackled multilingual generation-based dialogue systems.
%Previous research in domain adaptation has developed some feature-representation-transfer approaches \cite{daume2007frustratingly}, which are not suitable for multilingual tasks. \cite{Mrk2017Semantic} reply on high-quality cross-lingual resources to facilitate language transfer in domain-specific dialogue systems. No open-domain multilingual dialogue generation system has been built.
%Many studies in speech recognition \cite{heigold2013multilingual} and text sentiment analysis \cite{lo2017multilingual} prove that multilingual systems can improve performance for scarce resource languages. 

%\begin{CJK}{UTF8}{gkai}
%\begin{table}[tbp] \small \centering  \tabcolsep 8pt
%\begin{tabular}{|m{3.6cm}|m{3.7cm}|}
%\hline
%\textbf{English} & \textbf{Chinese}  \\
%\hline
%- how are you today? & - 你今天好吗？\\
%- Not bad. How about you? & - 不错啊。你呢？\\
%- Pretty good. & - 我很好。\\
%- That is great.& - 那太好了。 \\
%\hline
%- happy birthday! Hope you're having a great day. & - 生日快乐！希望你度过了愉快的一天。\\
%- thank you! Sorry for the late reply. & - 谢谢你！抱歉回复晚了。\\
%\hline
%
%\end{tabular}
%\caption{Examples of similar dialogues in different languages. The English and Chinese utterances are from Twitter and Weibo datasets, respectively.}
%\label{patterneg}
%\end{table}
%\end{CJK}
This paper proposes a multi-task learning architecture for multilingual open-domain dialogue system that leverages the common dialogue features shared among different languages. Inspired by \cite{miller2016key-value}, we augment the $\mathtt{Seq2Seq}$ framework by adding a architecture-improved key-value memory layer between the encoder and decoder. Concretely, the memory layer consists of two parts, where the key memory is used for query addressing and the value memory stores the semantic representation of the corresponding response. To capture both shared and private features in different languages, the memory layer is further divided into shared and private memory separately.
Though proposed for open-domain dialogue system, the multilingual shared-private memory architecture can be adapted flexibly and used for other tasks.

%In this paper, we build a multilingual open-domain dialogue system to leverage the common features shared among different languages. To capture the correspondence features between conversation query and response, we augment the $\mathtt{Seq2Seq}$ framework by adding a architecture-improved key-value memory layer between the encoder and decoders. Concretely, the memory layer consists of two parts, where the input memory is used for query addressing and the output memory stores the semantic representation of the corresponding response. To capture both shared and private features in different languages, the memory layer is further divided into shared and private memory separately.Though proposed for open-domain dialogue system, the multilingual shared-private memory architecture can be adapted flexibly and used for other tasks.
% Furthermore, the memory layer contains shared and private memory to capture both shared and private features in different languages.
% Due to the flexible design, the improved key-value memory network, the memory augmented Seq2Seq and the multi-task framework with shared-private memory can also be used for other tasks. 
Experiments conducted on Weibo and Twitter conversational corpora of different sizes show that our proposed multilingual architecture outperforms existing techniques on both automatic and human evaluation metrics. Especially when the training data is scarce, the dialogue capability can be enhanced significantly with the help of the multilingual model.

To this end, the main contributions of our work are summarized into four folds:
1) To the best of our knowledge, the proposed work is the first to provide a solution for multilingual dialogue systems.
2) We improve the traditional key-value memory structure to expand its capacity, with which we extend the $\mathtt{Seq2Seq}$ model to capture dialogue features.
3) Based on the memory augmented dialogue model, a multi-task learning architecture with shared-private memory is proposed to achieve the transfer of dialogue features among different languages. 
4) We empirically demonstrate the efficiency of multi-task learning in dialogue generation task and investigate some characteristics of this framework.

\section{Related Works}
\subsection{Dialogue Systems}
Building a dialogue system is a challenging task in natural language processing (NLP). The focus in previous decades was on template-based models \cite{Wallace2008The}. 
However, recent generation-based dialogue systems are of growing interest due to their effectiveness and scalability. 
%a large mount of data available.
Ritter et al. \cite{Ritter2011Data} proposed a response generation model using statistical machine-translation methods. This idea was further developed by \cite{sordoni2015a}, who represented previous utterances as a context vector and incorporated the context vector into response generation.
%Attention is another important part in dialogue generation.
Many methods are applied in dialogue generation. Attention helps the generation-based dialogue system by aligning the context and the response.
%\cite{Yao2015Attention} added an extra recurrent neural network (RNN) with attention mechanism to represent the structural process of intentions. 
\cite{mei2016coherent} improved the performance of a recurrent neural network dialogue model via a dynamic attention mechanism.
In addition, some works concentrate on many aspects of the dialogue generaion, including diversity, coherence, personality, knowledgeable and controllability \cite{yan2018chitty}. In these approaches, the corpora used are always in the same language. These systems are referred to as monolingual dialogue systems. As far as we know, this study is the first to explore the use of multilingual architecture to better suit the generation-based dialogue system.

\subsection{Memory Networks}
Memory networks \cite{2014arXiv1410.3916W,sukhbaatar2015end-to-end} are a class of neural network models that are augmented with external memory resources. Valuable information can be stored and reused in memory networks through the memory components. Based on the end-to-end memory network
architecture \cite{sukhbaatar2015end-to-end}, \cite{miller2016key-value} proposed a key-value memory network architecture for question answering. The memory stores facts in a key-value structure so that the model can learn to use keys to address relevant memories with respect to the question and return corresponding values for answering. \cite{bordes2016learning}, \cite{Madotto2018Mem2Seq} and \cite{Wu2018End} built goal-oriented dialogue systems based on memory-augmented neural networks. Compared with the above models, our memory components are not trained based on specific knowledge bases, but self-tuning in the training process, which makes the model more flexible. We further divide each memory module into several blocks to improve its capability.

\subsection{Multi-task learning}
Multi-task learning (MTL) is an approach to learn multiple related tasks simultaneously. It improves generalization by leveraging the domain-specific information contained in the training signals of related tasks \cite{caruana1998multitask}. \cite{Collobert2008A} confirmed that NLP models benefit from the MTL approach. Many recent deep-learning approaches to multilingual issues also used MTL as part of their model. 

In the context of deep learning, MTL is usually done with either hard or soft parameter sharing of hidden layers: hard parameter sharing method explicitly shares hidden layers between tasks while keeping several task-specific output layers; soft parameter sharing method usually employs regularization techniques to encourage the parameters in different tasks to be similar \cite{Ruder2017An}. Hard parameter sharing is the most commonly used approach to MTL in neural networks. \cite{Dong2015Multi} learned a model that simultaneously translated sentences from one source language to multiple target languages. \cite{liu2017adversarial} propose an adversarial multi-task learning framework for text classification. \cite{2017arXiv170605137K} demonstrated a single deep learning model that jointly learned large-scale tasks from various domains including multiple translation tasks, an English parsing task, and an image captioning task. However to date, no multilingual dialogue-generation system based on multi-task learning framework has been built.

\section{Model}
%In this section, we first review the vanilla $\mathtt{Seq2Seq}$ dialogue system. Then, we introduce the $\mathtt{MemSeq2Seq}$ model which adds a key-value memory layer between the encoder and decoder to learn dialogue features, and the $\mathtt{ImpMemSeq2Seq}$ which divides the memory of $\mathtt{MemSeq2Seq}$ into blocks to expand model capacity. The models mentioned above are the basis of our multilingual model which are further extended with shared-private memory components to implement the multilingual dialogue systems.
In this section, we first review the vanilla $\mathtt{Seq2Seq}$, then propose the key-value memory augmented $\mathtt{Seq2Seq}$ models, and extended them with shared-private memory components to implement the multilingual dialogue systems.

\subsection{Preliminary background knowledge} \label{sec:Preliminary}
A $\mathtt{Seq2Seq}$ model maps input sequences to output sequences. It consists of two key components: an encoder, which encodes the source input to a fix-sized context vector using the Recurrent Neural Network (RNN), and a decoder, which generates the output sequence with another RNN based on the context vector. 

Given a source sequence of words (query) $q=\{x_1,x_2,...,x_{n_q}\}$ and a target sequence of words (response) $r=\{y_1,y_2,...,y_{n_r}\}$, a basic $\mathtt{Seq2Seq}$ based dialogue system automatically generates response $r$ conditioned on query $q$ by maximizing the generation probability $p(r|q)$. Specifically, the encoder encodes $q$ to a context vector $c$, and the decoder generates $r$ word by word with $c$ as input. The objective function of $\mathtt{Seq2Seq}$ can be written as
\begin{eqnarray}
h_t=\mathtt{f}(x_t,h_{t-1}),~c=h_{n_q}, ~~~~~~\\
p(r|q)=p(y_1|c)\prod_{t=2}^{n_r}p(y_t|c,y_1,...,y_{t-1}),
\end{eqnarray}
where $h_t$ is the hidden state at time $t$ and $f$ is a non-linear transformation. Moreover, gated recurrent units (GRU) and the attention mechanism proposed by \cite{bahdanau2015neural} are used in this work. 
% All of the following proposed models are extended base this model.

%Compare Imp-Mem-Seq2Seq with Mem-Seq2Seq in detail, Mem-Seq2Seq's memory contains only one block, in which the size of I and O are both $|V|\times t$, and the output $U$ is a weighted combination of $t$ entries. Imp-Mem-Seq2Seq's memory contains $n$ blocks, the size of $I_i$ and $O_i$ in each block are both $|V|/n \times t$, and the output of each block $U_i$ is also a weighted combination of $t$ entries. But key addressing and value calculation in $n$ blocks are done independently, and the output of Imp-Mem-Seq2Seq's memory is a combination of $U_1, U_2, \ldots, U_n$, so the capacity is significantly improved.

\begin{figure*}[tbp]
\centering
\includegraphics[width=.9\linewidth]{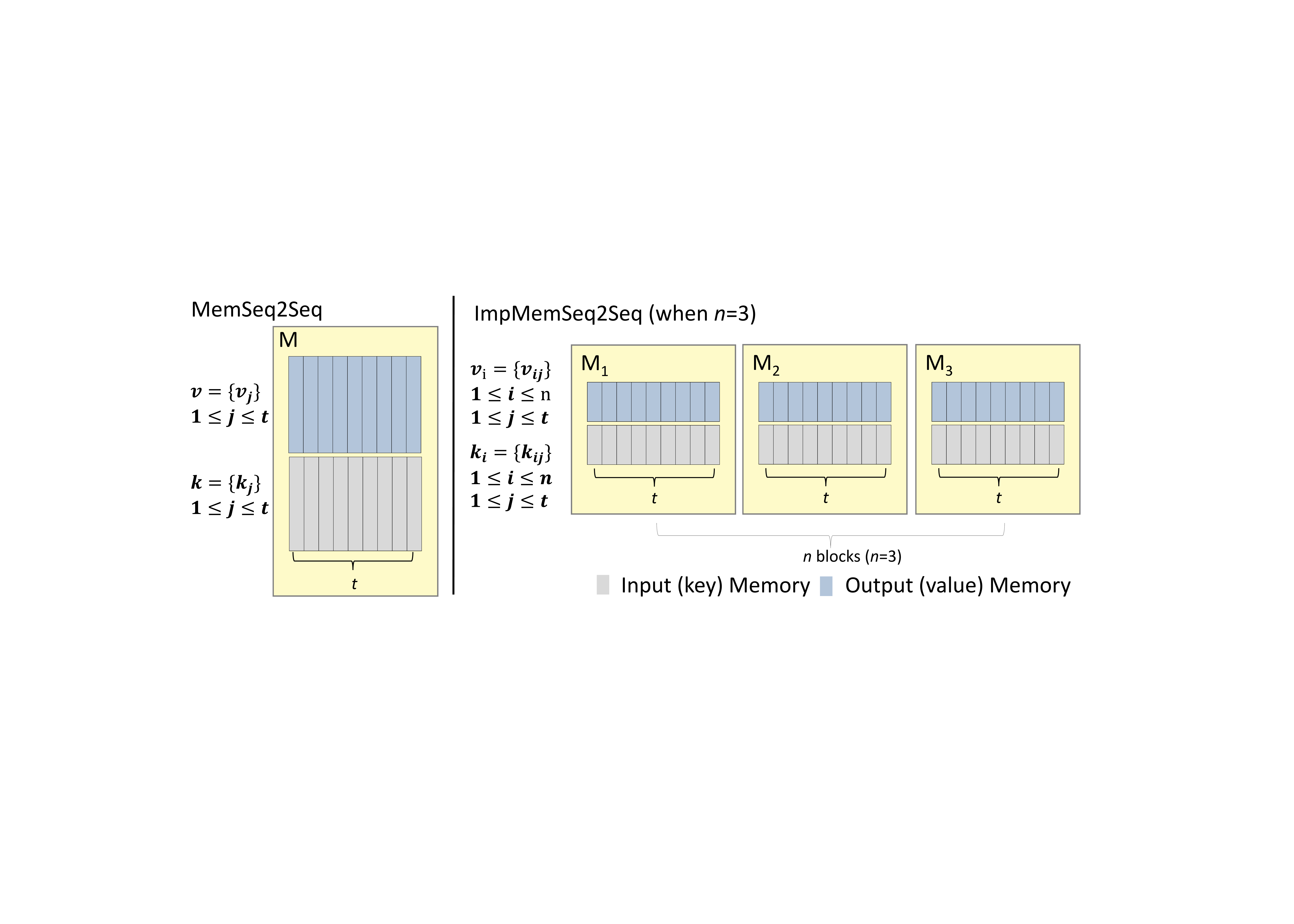}
\caption{Memory structure of $\mathtt{MemSeq2Seq}$ and $\mathtt{ImpMemSeq2Seq}$. $M$ represents memory module. $k$ and $v$ represent input and output memory respectively.  }
\label{fig_memory}
\end{figure*}

\subsection{Key-Value Memory Augmented $\mathtt{Seq2Seq}$}

Inspired by the end-to-end memory network\cite{miller2016key-value}, we introduce the $\mathtt{MemSeq2Seq}$ model which adds a key-value memory layer between the encoder and decoder to learn dialogue features, and the $\mathtt{ImpMemSeq2Seq}$ which divides the memory of $\mathtt{MemSeq2Seq}$ into blocks to expand model capacity.

%Specifically, the memory is randomly initialized and jointly learned during the training process without an explicit knowledge base.

\subsubsection{MemSeq2Seq}
\label{sec:Memory} The $\mathtt{MemSeq2Seq}$ augments the $\mathtt{Seq2Seq}$ with a key-value memory layer between the encoder and decoder. The memory component consists of two parts: input (key) and output (value) memorys.
%(see the left of Figure \ref{fig_memory}) 
The input memory is used for query representation addressing, while the output memory stores the representation of the corresponding response information. 
%It should be noted that correspondence in dialogue is simplified here, each query has only one corresponding answer. Therefore, 
%The number of key-value pairs is the number of slots in the memory. 
The model retrieves information from the value memory with the weights computed as the similarity between the query representation and the key memory, with the goal of selecting values that are most relevant to the query.
% Thus, the model can help calibrate the selection of the first word to make it more accurate, and the first word has a beneficial effect on the subsequent decoding results.

Formally, we first encode a query $q$ to a context vector $c$, and then calculate the similarity $p=\{p_1,...p_t\}$ between $c$ and each item of the key memory using softmax weight. Later, the model computes a new context vector $c^*$, which is a weighted sum of the value memory according to $p$.
%the model computes $u$ by
%decoder input vector $U$ is a weighted sum of output memories $O$ according to $P$, where the dimensions of $V$ and $U$ are identical. 
%Below is the calculation process.
\begin{eqnarray}
\setlength{\abovedisplayskip}{1pt}
\setlength{\belowdisplayskip}{1pt}
p_{j}=\mathtt{softmax}(c\cdot k_{j}),~~  \\
   c^*=\sum_{j=1}^t p_{j}v_{j}~~(1\leq j\leq t),  
\end{eqnarray}
where $k_{j}$ and $v_{j}$ are items in the key and value memory, and $t$ is the number of key and value items. During training, all items in memory and parameters in the $\mathtt{Seq2Seq}$ are jointly learned to maximize the likelihood of generating the ground-truth responses conditioned on the queries in the training set.

\subsubsection{ImpMemSeq2Seq}\label{sec:Improved}
In the $\mathtt{MemSeq2Seq}$, the key-value pairs in memory are limited, which are linear with the number of items in memory. 
To expand capacity, we further divide the entire memory into several individual blocks and accordingly split the input vector into several segments
to compute the similarity scores.
After division, similarity to multi-head attention mechanism~\cite{vaswani2017attention}, different representation subspaces at different positions are individually projected and the number of key-value pairs becomes the number of slot combinations in these blocks, while one key still corresponds to one value. 

The model first split a context vector $c$ into $n$ segments, then compute new context segments $c^*_i$ by memory blocks independently, and the final new context vector $c^*$ is the concatenation of $c^*_i$. The formula is as follows. 
\begin{eqnarray}
  c_1,...c_n = \mathtt{split}(c), ~~
  c^*_i = \mathtt{M_i}(c_i), ~~~~~\\
  c^*=\mathtt{concat}(c^*_1,c^*_2,\ldots,c^*_n), ~~~~~~~~~~
\end{eqnarray}
where $M_i$ represents the calculation in $i^{th}$ memory block.

The $\mathtt{ImpMemSeq2Seq}$ calculates the weight $p$ with a finer granularity, which makes the addressing more precise and flexible. Besides, with a parallel implementation, the memory layer becomes more efficient.

\begin{figure*}[ht]
\centering
\includegraphics[width=1\linewidth]{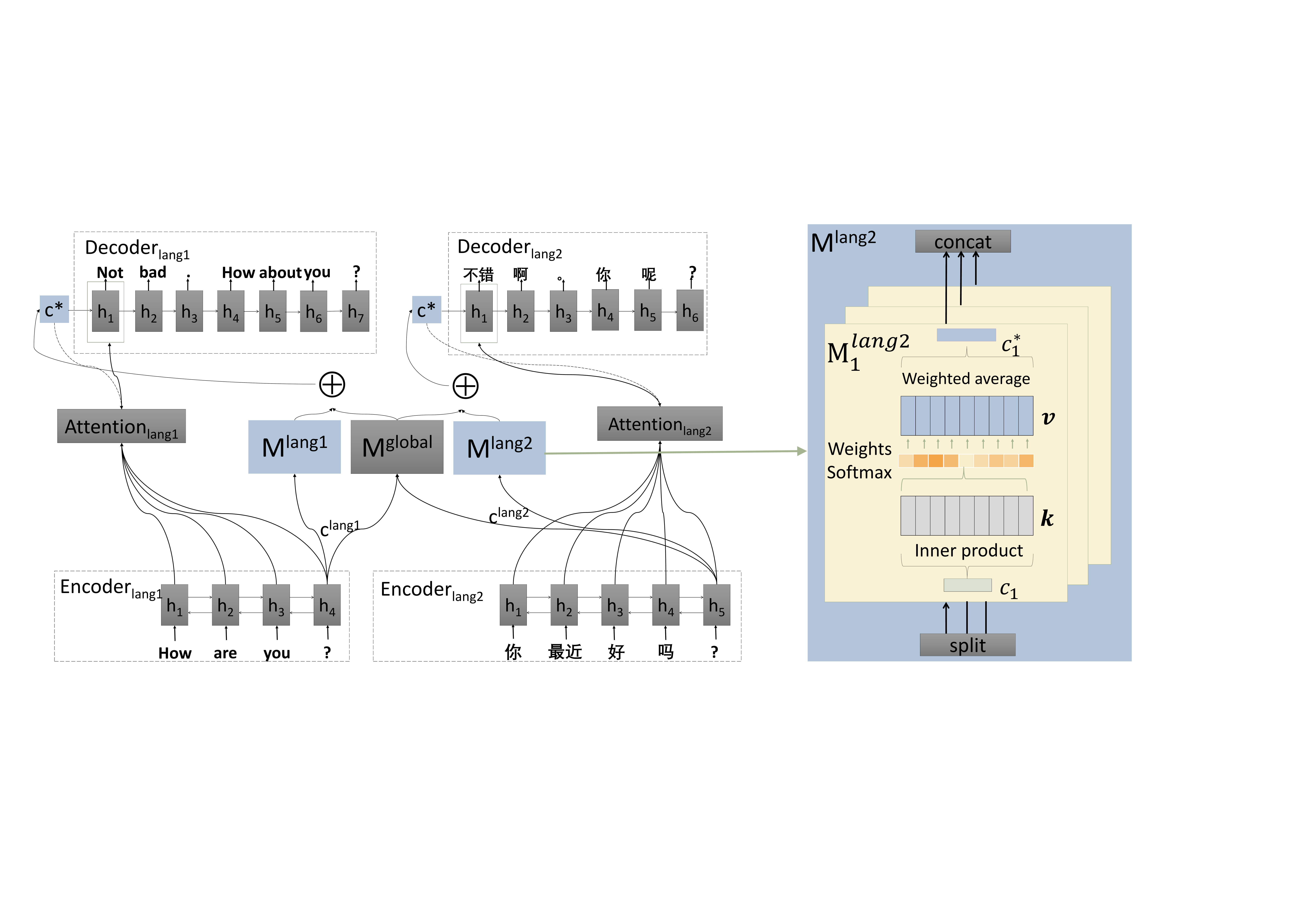}
\caption{$\mathtt{SPImpMem}$ model for multilingual dialogue system.   $M^{global}$ represents the shared memory. Superscript \textit{lang1}, \textit{lang2} represent two different languages respectively.}
\label{fig_share_memory}
\end{figure*}
\subsection{$\mathtt{Seq2Seq}$ with Shared-Private Memory}
\label{sec:Shared-private}
The models introduced in the previous sections can be extended for monolingual tasks. Specifically, we augment the $\mathtt{MemSeq2Seq}$
and $\mathtt{ImpMemSeq2Seq}$ for multilingual tasks and named the extensions $\mathtt{SPMem}$
and $\mathtt{SPImpMem}$, respectively.
According to multi-task learning, dialogue systems in two different languages can be simultaneously trained. By sharing representations between two dialogue tasks, the model facilitates the cross-lingual transfer of dialogue capability.

Our multilingual model consists of four modules: an encoder, decoders, a private memory for each language and shared memory occupied by all languages. Figure \ref{fig_share_memory} gives an illustration of $\mathtt{SPImpMem}$. The $\mathtt{SPMem}$ is a special case where the number of memory blocks $n$ is set to 1. More specifically, given a input query $q$, the encoder of its language first encode it into a context vector $c$, and then the model feeds $c$ to both its private and shared memory. The private memory is occupied by the language corresponding to the input. The shared memory is expected to capture common features of conversations among different languages. By matching and addressing the shared and private memory components, we obtain two output vectors that are then concatenated as a new context vector $c^*$. The returned vector is supposed to contain features from both its own language and other languages involved in the multilingual model, which is then fed to the decoder of its language.

%\textbf{Objective Function}: 

Given the first language conversational corpus $(\textbf{q}_{i}^{1},\ \textbf{r}_{i}^{1})_{i=1}^{T_1}$ and the second language conversational corpus $(\textbf{q}_{i}^{2},\ \textbf{r}_{i}^{2})_{i=1}^{T_1}$, the parameters $\Theta$ are learned by minimizing the negative log-likelihood between the generated $\widetilde{r}$ and reference \textbf{r}, that is equivalent to maximizing the conditional probability of responses $\textbf{r}_{1}$ and $\textbf{r}_{2}$ given  $\Theta$, $\textbf{q}_{1}$ and $\textbf{q}_{2}$:
\begin{eqnarray}
J=\frac{1}{T_1}\sum_{i=1}^{T_1}\log p(\textbf{r}_{i}^{1}|\textbf{q}_{i}^{1},\Theta_{s_1},\Theta_{M_1},\Theta_{M_g})  
+\frac{1}{T_{2}}\sum_{i=1}^{T_{2}}\log p(\textbf{r}_{i}^{2}|\textbf{q}_{i}^{2},\Theta_{s_2},\Theta_{M_2},\Theta_{M_g}),
\end{eqnarray}
where $\Theta_{S}$ is a collection of parameters for the encoders and decoders; $\Theta_M$ is the parameters of memory contents; $T$ is the size of corpus; and subscriptions \textit{1}, \textit{2} and \textit{g} represent \textit{lang1}, \textit{lang2}, and \textit{global} in Figure 2 respectively.

\section{Experimental Settings}
\subsection{Datasets}
%Our model can be extended to any number of languages: without loss of generality, 
We conducted experiments on open-domain single-turn Chinese (Zh) and English (En) conversational corpora% \footnote{The model can be extended to any number of languages.}
. The Chinese corpus consists of 4.4 million conversations and the English corpus consists of 2.1 million conversations \cite{Ritter2011Data}. The conversations are scraped from Sina Weibo\footnote{http://weibo.com\label{weibo}} and Twitter\footnote{http://www.twitter.com} respectively. 
%Short and meaningless replies are filtered. to do by zhenxinfu 
%Some examples are given in Table \ref{patterneg}. 
%However, a lot of query-response pairs in the Twitter corpus are not dialogues. 
%The quality of the Weibo corpus is much better than that of the Twitter. One reason is that the Twitter data contains many casual words, which leads to the difference in experimental performances.

The experiments include two parts: balanced and unbalanced tests, which are discriminated by the relative size of training data for each language. In the balanced tests, the sizes of the Chinese and English corpus are comparable. 
We empirically set the dataset size to 100k, 400k, 1m and the whole (4.4m-Zh, 2.1m-En) to evaluate the model performance in different data scales. The unbalanced tests consist of training data of (1m-Zh, 100k-En) and (100k-Zh, 1m-En) respectively. Subsets used are sampled randomly. All the experiments have the same validation and testing data with size 10k.
%The dictionary size of the open domain dialogue system is usually set to tens of thousands. Therefore, use 100k dialogue pairs training dialogue model is low-resource, and most related research is based on millions corpus size \cite{song2016two,wu2018neural,wang2018chat}.

\subsection{Evaluation Metrics}
Three different metrics are used in our experiments:%, containing  automatic metrics and human evaluation:
%\begin{itemize}
%\setlength{\itemsep}{0pt}
%\setlength{\parsep}{0pt}
%\setlength{\parskip}{0pt}

\begin{itemize}
    \item \textbf{Word overlap based metric}: Following previous work \cite{sordoni2015a}, we employ BLEU \cite{papineni2002bleu} as an evaluation metric to measure word overlaps in a given response compared to a reference response. 
    \item \textbf{Distinct-1 \& Distinct-2}: Distinct-1 and Distinct-2 are the ratios of distinct unigrams and bigrams in generated responses respectively \cite{li2015diversity} which measure the diversity of the generated responses.
    \item \textbf{Three-scale human annotation}: We adopt human evaluation following \cite{wu2018neural}. Four human annotators were recruited to judge the quality of 500 generated responses from different models. All of the responses are pooled and randomly permuted. The criteria are as follows: \textbf{+2}: the response is relevant and natural; \textbf{+1}: the response is a correct reply, but contains little errors; \textbf{0}: the response is irrelevant, meaningless, or has serious grammatical errors.
\end{itemize}

%\end{itemize}

\subsection{Implementation Details}
%%The system is implemented in PyTorch\footnote{http://pytorch.org/}. 
The Adam algorithm is adopted for optimization during training. All embeddings are set to 630-dimensional and hidden states 1024d. Considering both efficiency and memory size, we restrict both the source and target vocabulary to 60k and the batch size to 32. Chinese word segmentation is performed on Chinese conversational data.
%\footnote{https://pypi.python.org/pypi/jieba/} 
%%English word tokenization is implemented based on NLTK.\footnote{http://www.nltk.org} 
For the single block memory components, the number of cells in the memory block is set to 1024 empirically, and the dimension of each cell is adjusted according to the encoder. The memory block is further divided into 32 parts in our improved memory model. In multilingual models, the number of blocks for shared and private memory component are the same. To prevent the multilingual model from favoring one certain language, we switched sentences of different languages individually by batch during training.

\subsection{Comparisons}
%The competing methods in experiments are listed below.
We compare our framework with the following methods:
%\begin{itemize}
%\setlength{\itemsep}{0pt}
%\setlength{\parsep}{0pt}
%\setlength{\parskip}{0pt}
\begin{itemize}
    \item \textbf{$\mathtt{Seq2Seq}$}. A $\mathtt{Seq2Seq}$ model with attention mechanism.
    \item \textbf{$\mathtt{MemSeq2Seq}$}. The key-value memory augmented $\mathtt{Seq2Seq}$ model in Section~\ref{sec:Memory}.
     \item\textbf{$\mathtt{ImpMemSeq2Seq}$}. The improved memory augmented $\mathtt{Seq2Seq}$ model with the memory block decomposed into several blocks as in Section~\ref{sec:Improved}.
    \item \textbf{$\mathtt{SPMem}$}. The proposed multilingual model with shared-private memory which is extended from $\mathtt{MemSeq2Seq}$. It is a special case of $\mathtt{SPImpMem}$ where the number of memory blocks \textbf{\textit{n}} is set to 1.
     \item \textbf{$\mathtt{SPImpMem}$}. The proposed multilingual model with shared-private memory component which is extended from $\mathtt{ImpMemSeq2Seq}$ as in Section ~\ref{sec:Shared-private}.
\end{itemize}

\section{Results and Analysis}
We present the evaluation results of balanced test and unbalanced test in Table \ref{table1} and \ref{tab2} respectively. Table \ref{table1} contains evaluation results of monolingual dialogue systems with Seq2Seq, $\mathtt{MemSeq2Seq}$ and $\mathtt{ImpMemSeq2Seq}$ as baseline. Table \ref{tab2} can be viewed in conjunction with the data in Table \ref{table1}. 
%Some cases are shown in the %Appendix\footnote{\url{https://anonymousfiles.io/vSovdLMZ/}} from test examples of low-resource datasets.
%Because the multilingual model transfers the cross-lingual knowledge, it generates texts of higher diversity and richer information. 
%Since models benefits a lot from the multilingual model when training data is scarce, Table \ref{table3} and \ref{table4} shows more detailed evaluation results of models trained with 100k dataset.

%Looking at the evaluation results in both Table \ref{table1} and Table \ref{tab2},

\subsection{Monolingual Models}
From Table \ref{table1}, we observe that the performance of the $\mathtt{MemSeq2Seq}$ model only slightly outperforms the $\mathtt{Seq2Seq}$ model. However, with memory decomposed into several parts, the $\mathtt{ImpMemSeq2Seq}$ model surpasses the basic $\mathtt{Seq2Seq}$ model. Therefore, we conclude from the comparisons that our modification of the memory components improves the capability of the model.
Another observation is that in English a good conversation model can be trained with less data. Hence it does not get a significant performance gain in English as the size of data increases.

%However, because of the differences in language characteristics and corpus quality, the Chinese and English experimental performance is not exactly the same. There is a big gain in Chinese from 100K to 400K but not for English.

%\begin{figure}[tbp]
%\centering
%\includegraphics[scale=0.35]{Balanced_Test_With_Multi.png}
%\caption{The curves of BLEU scores got by the proposed multilingual model shows a definite improvement when compared with the several monolingual baselines.}
%\label{balanced_test_with_multi}
%\end{figure}

\begin{table*}[ht]
\setlength{\abovecaptionskip}{0.3cm}
\setlength{\belowcaptionskip}{-0.1cm}
\centering
\scriptsize
\begin{tabular}{|cc|c|c|c|c|c|}
\hline
\multicolumn{2}{|c|}{\multirow{2}*{Datasets}}&\multicolumn{3}{c|}{Monolingual (baseline)} & \multicolumn{2}{c|}{Multilingual}\\ \cline{3-7}
 & & Seq2seq & $\mathtt{MemSeq2Seq}$ & $\mathtt{ImpMemSeq2Seq}$ & $\mathtt{SPMem}$ & $\mathtt{SPImpMem}$ \\ \hline
\multirow{2}*{100k} & \multicolumn{1}{|c|}{Zh} & 0.485 & 0.478 & 0.492 & 0.524 & {\bf 0.549} \\ \cline{2-7}
& \multicolumn{1}{|c|}{En} & 0.779 & 0.780 & 0.825 & {\bf0.831} & 0.805 \\ \hline
\multirow{2}*{400k} & \multicolumn{1}{|c|}{Zh} & 2.024 & 2.144 & 2.142 & {\bf 2.565} & 2.317 \\ \cline{2-7}
& \multicolumn{1}{|c|}{En} & 1.001 & 0.937 & 0.974 & 0.976 & {\bf1.034} \\ \hline
\multirow{2}*{1m} & \multicolumn{1}{|c|}{Zh} & 3.135 & 3.131 & {\bf3.268} & 2.732 & 2.827 \\ \cline{2-7}
& \multicolumn{1}{|c|}{En} & 1.027 & 1.100 & 1.099 & 1.135 &{\bf1.146} \\ \hline
\multirow{2}*{all (4.4m-Zh, 2.1m-En)} & \multicolumn{1}{|c|}{Zh} & 3.600 & 3.383 & {\bf 3.755} & 2.955 & 2.765 \\ \cline{2-7}
& \multicolumn{1}{|c|}{En} & 1.082 & 1.140 & 1.208 & 1.254 & {\bf1.336} \\ \hline
\end{tabular}
\caption{BLEU-4 scores of the {\bf balanced test}. The results of monolingual experiments are included for comparison.}
\label{table1}
\end{table*}
%, where the size of Chinese (Zh) and English (En) training data are comparable

\begin{table}[tbp]
\setlength{\abovecaptionskip}{0.3cm}
\setlength{\belowcaptionskip}{-0.1cm}
\centering 
\scriptsize
\tabcolsep 10pt
\begin{tabular}{|cc|c|c|}
\hline
\multicolumn{2}{|c|}{\multirow{2}*{Datasets}}& \multicolumn{2}{c|} {Multilingual}\\ \cline{3-4}
& &  $\mathtt{SPMem}$ & $\mathtt{SPImpMem}$\\ \hline
\multirow{2}*{100k-Zh, 1m-En} & \multicolumn{1}{|c|}{Zh}  & 1.442 & {\bf 1.484} \\ \cline{2-4}
& \multicolumn{1}{|c|}{En}  & {\bf 1.083} & 1.075 \\ \hline
\multirow{2}*{1m-Zh, 100k-En} & \multicolumn{1}{|c|}{Zh} & 2.607 & {\bf 2.690} \\ \cline{2-4}
& \multicolumn{1}{|c|}{En} &  0.800 & {\bf 0.839} \\ \hline
\end{tabular}
\caption{BLEU-4 scores of the {\bf unbalanced test}. Numbers in bold mean that it achieves the best performance among all models trained with this dataset.}
\label{tab2}
\end{table}

\subsection{Multilingual Models}
\subsubsection{Balanced Test}
From the experimental results shown in Table \ref{table1}, we observe that the proposed multilingual model outperforms the monolingual baselines on English corpus of different sizes. For the Chinese corpus, the promotion decreases when the size of training data increases, and thus it can only be seen on data of small sizes (i.e., 100k and 400k). Similar results can also be observed in \cite{firat2016multi-way}. There are several interpretations of the phenomena: 
1) By the shared memory component in the proposed multilingual model, common features are learned and transferred through both languages. Thus, when one language corpus is insufficient, some common features from other languages are helpful.
2) With the scale of corpus increasing, the monolingual model is already capable enough so that noisy information from other languages may hinder the original system.

Nevertheless, the contrary behaviors of the multilingual model on Chinese and English corpus remain suspended. As the scale of training data grows, the performance of $\mathtt{SPMem}$ and $\mathtt{SPImpMem}$ on English corpus outperforms the monolingual baselines while the performance decreases on Chinese corpus.
This may result from the various qualities of different corpora which further influence the features in the shared memory blocks. 
The Chinese monolingual model whose parameters are originally well estimated are hindered by the noise from the shared memory. However, the English monolingual model that is relatively poorly trained benefit from the shared features. In a word, the higher quality corpus needs multilingual training less. Our model focuses on the scenario that the corpus of one language is scarce.

%{\bf \noindent Unbalanced Test }
\subsubsection{Unbalanced Test}
Since models benefit a lot from the multilingual model when training data is scarce in Table \ref{tab2}, we present more detailed evaluation results of models trained with the 100k datasets in Table \ref{table4} and Table \ref{table3}. % and the results for Chinese dataset is shown in Appendix\footnote{\url{https://anonymousfiles.io/vSovdLMZ/}} for page limitation. 
It is clear that, with the help of another rich resource language corpus, the multilingual model improves the performance of language with limited training data on automatic evaluation metrics except for Distinct-1. The improvements remain true even when comparing the unbalanced test results with the balanced test results, which are strengthened by the other language corpus with the same size. According to the human evaluation results, $\mathtt{SPMem}$ and $\mathtt{SPImpMem}$ generate more informative and interesting responses (+2 responses) but perform much worse on +1 responses for grammatical errors. Fleiss' Kappa on all models are larger than 0.4,  which proves the correlation of the human evaluation. Therefore, some features captured by the shared memory from one language can be efficiently utilized by other languages.

\begin{table*}[tbp]
\setlength{\abovecaptionskip}{0.3cm}
\setlength{\belowcaptionskip}{-0.1cm}
\centering \scriptsize 
\resizebox{\textwidth}{!}{
\begin{tabular}{|l|cccc|cc|cccc|}
\hline
Dataset (100k-En) & BLEU-1 & BLEU-2 & BLEU-3 & BLEU-4 & Distinct-1& Distinct-2 &0&+1&+2&Kappa\\  \hline
$\mathtt{Seq2Seq}$&8.513&3.331&1.566&0.779&0.019&0.077&0.36&0.60&0.04& 0.54\\ \hline
$\mathtt{MemSeq2Seq}$&8.613&3.378&1.576&0.780&0.017&0.061&0.38&0.59&0.03& 0.58\\ 	
$\mathtt{ImpMemSeq2Seq}$&9.178&3.623&1.682&0.825&0.018&0.070&0.37&0.58&0.05& 0.47\\
$\mathtt{SPMem}$ (with 100k-zh)&8.844&3.520&1.651&0.831&0.017&0.071&0.56&0.41&0.03& 0.43 \\	
$\mathtt{SPImpMem}$ (with 100k-zh)&9.048&3.573&1.637&0.805&0.020&0.080&0.52&0.44&0.04& 0.58 \\ \hline
$\mathtt{SPMem}$ (with 1m-zh)&\textbf{10.669}&\textbf{3.686}&1.580&0.800&0.007&\textbf{0.124}&0.31&0.62&0.07 & 0.46\\
$\mathtt{SPImpMem}$ (with 1m-zh)&9.118&3.648&\textbf{1.701}&\textbf{0.839}&\textbf{0.025}&0.103&0.52&0.33&0.15&0.51\\ \hline
\end{tabular}}
\caption{Evaluation results of models trained with the 100k English dataset. }
\label{table3}
\end{table*}

\begin{table*}[tbp]
\setlength{\abovecaptionskip}{0.3cm}
\setlength{\belowcaptionskip}{-0.1cm}
\centering \scriptsize 
\resizebox{\textwidth}{!}{
\begin{tabular}{|l|cccc|cc|cccc|}
\hline
Dataset(100k-Zh) & BLEU-1 & BLEU-2 & BLEU-3 & BLEU-4 & Distinct-1& 
Distinct-2 &0&+1&+2&Kappa\\ 
\hline
$\mathtt{Seq2Seq}$ & 9.463 & 3.035 & 1.168 & 0.485 & \textbf{0.026} & 0.113 & 0.47 & 0.45 & 0.08 & 0.74 \\
\hline
$\mathtt{MemSeq2Seq}$&9.210&2.859&1.101&0.478&0.018&0.073&0.43&0.43 &0.14&0.74 \\
$\mathtt{ImpMemSeq2Seq}$&9.682&3.041&1.164&0.492&0.021&0.094&0.54 &0.34 	&0.12&0.65 \\
$\mathtt{SPMem}$(with 100k-en)&10.329&3.132&1.213&0.524&\textbf{0.026}&0.114&0.52&0.33&0.15&0.70 \\
$\mathtt{SPImpMem}$(with 100k-en)&9.978&3.136&1.242&0.549&\textbf{0.026}&0.114&0.32&\textbf{0.53}&0.15&0.76 \\
\hline
$\mathtt{SPMem}$(with 1m-en)&11.940&\textbf{4.193}&2.211&1.442&0.017&0.148&
\textbf{0.55}&0.27&0.18&0.82 \\
$\mathtt{SPImpMem}$(with 1m-en)&\textbf{11.991}&4.179&\textbf{2.236}&\textbf{1.484}&0.019&\textbf{0.164}&0.53&0.23&\textbf{0.24}&0.85 \\
\hline
\end{tabular}}
\caption{Evaluation results of models trained with the 100k Chinese dataset. }
\label{table4}
\end{table*}

\begin{figure}[!tp]
\centering
\includegraphics[scale=0.23]{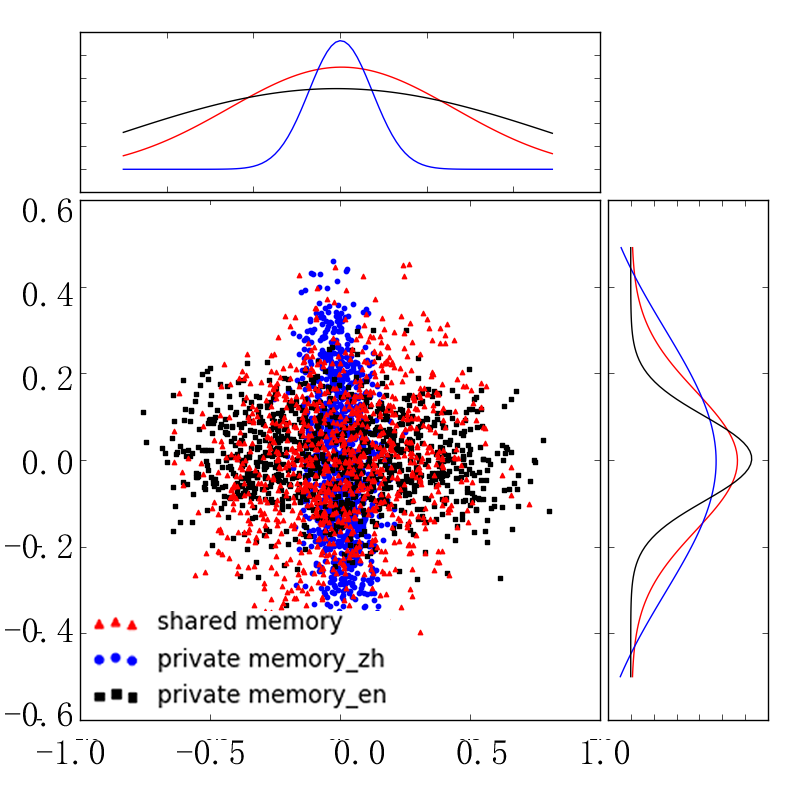}
\caption{The figure shows a 2-dimensional PCA projection of the input block in the memory networks. The two private memory blocks are differently oriented, and the shared memory block tends to be the mixture of them. The curves located above and right show more details of the distributions along two axes.}
\label{memory_show}
\end{figure}

\subsection{Model Analysis} 
To illustrate the information stored in the memory components, Figure \ref{memory_show} visualizes the first input block of each memory, namely two private and one shared memory components. From the scatter diagram and the fitting results of the Gaussian distribution, we observe some characters in the memory layer. Tuned explicitly by each separate language, the two private memory blocks learn and store different features that appear to distribute differently in the two dimensions after principal component analysis (PCA) projection. Nevertheless, the shared memory that is jointly updated by the two languages is likely to keep some common features of each private memory block. % and thus it is a blend of the private memories.

%Our model can also be extended to more languages theoretically. However, it's not appropriate to share a memory component among more languages, because larger datasets with more languages may introduce more noise. The methods of shared memory should be improved accordingly.

\section{Conclusion}
%This paper proposes a neural network architecture for multilingual open-domain dialogue systems. Based on the multi-task $\mathtt{Seq2Seq}$ model with private-shared memory components, the proposed architecture improves dialogue generation performance by exploiting non-parallel corpora. The private memory is occupied by each separate language, and the shared memory is expected to capture and transfer common dialogue features among different languages. To expand the capacity of vanilla memory network, the entire memory is further divided into individual blocks. Abundant experimental results show that our model outperforms individually learned monolingual models when the training data is limited.

This paper proposes a multi-task learning architecture with share-private memory for multilingual open-domain dialogue generation. The private memory is occupied by each separate language, and the shared memory is expected to capture and transfer common dialogue features among different languages by exploiting non-parallel corpora. To expand the capacity of vanilla memory network, the entire memory is further divided into individual blocks. Experimental results show that our model outperforms separately learned monolingual models when the training data is limited.

%As the size of the training data increases, we also observe different experimental performances for different languages. We ascribe this to that datasets with different quanlities may lead to different parameter estimations which will further incur the discrepancies among different models. A future research direction could be efficiently scheduling the transferring of features among multiple languages to help one certain language.

 \section*{Acknowledgments}
 \label{sec:Acknowledgments} 
We thank the anonymous reviewers for their insightful comments on this paper.
This work was supported by the National Key Research and Development Program of China (No. 2017YFC0804001), the National Science Foundation of China (NSFC No. 61876196 and 61672058). %The corresponding author is Rui Yan. %Rui Yan was sponsored by Microsoft Research Asia (MSRA) Collaborative Research Program.

\end{document}